\documentclass[10pt,twocolumn,letterpaper]{article}

\usepackage{cvpr}              

\usepackage{graphicx}
\usepackage{multirow}
\usepackage[most]{tcolorbox}
\usepackage{amssymb}
\usepackage[table]{xcolor}
\usepackage{tabularx}

\definecolor{mainquestionbg}{RGB}{255, 245, 245} 
\definecolor{mainquestionframe}{RGB}{200, 80, 80}  
\usepackage{enumitem}
\usepackage{bm}

\definecolor{cvprblue}{rgb}{0.21,0.49,0.74}
\PassOptionsToPackage{hyperfootnotes=false}{hyperref}
\usepackage[breaklinks,colorlinks,allcolors=cvprblue]{hyperref}
\usepackage[hang,flushmargin]{footmisc}

\makeatletter
\apptocmd{\@maketitle}{%
  \vspace{-1em}
\begin{center}
    \centering
    \includegraphics[page=1,width=1.0\textwidth,trim=0 0 0 24pt,clip]{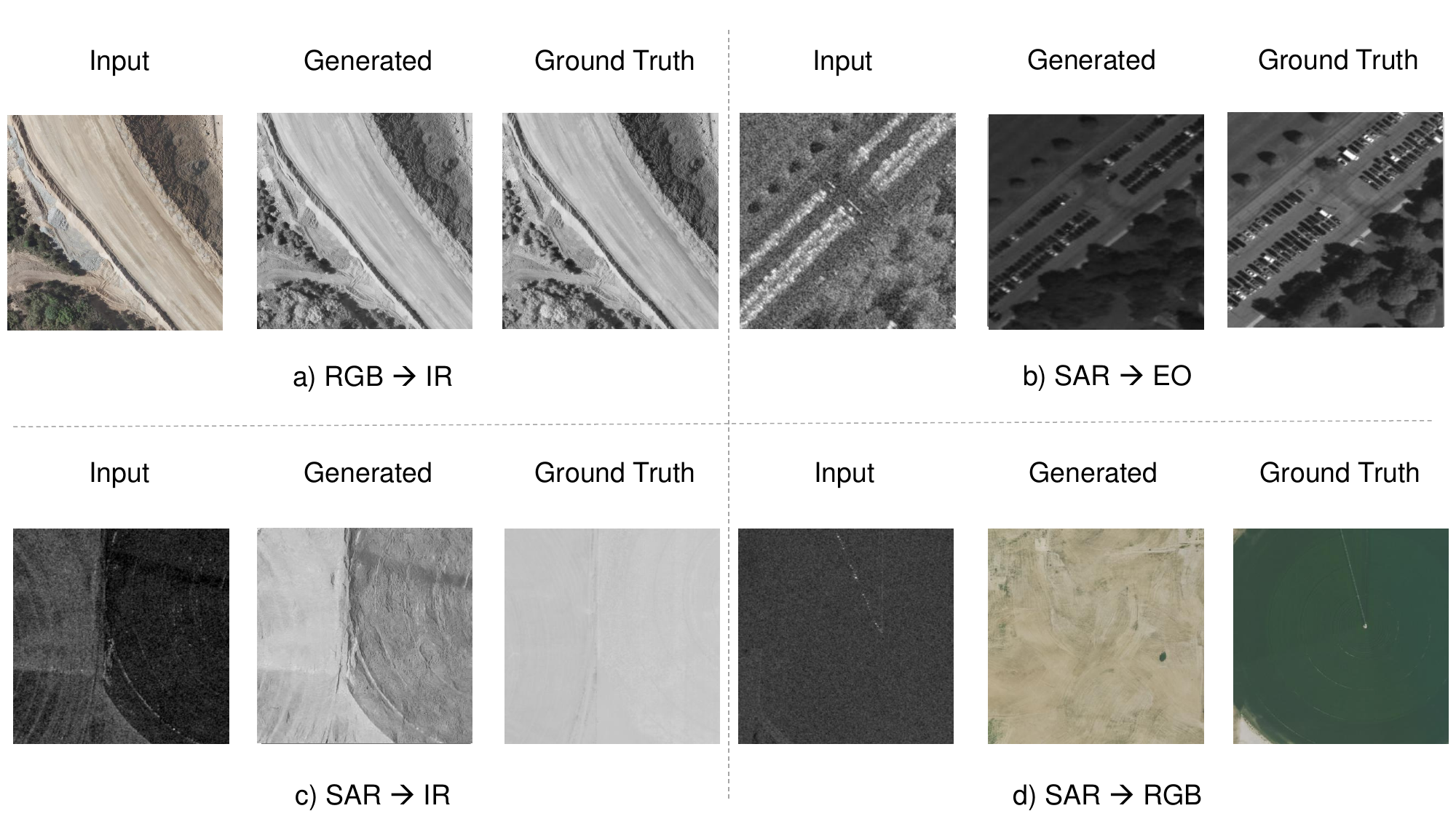}
    \captionof{figure}{Qualitative results on our hold-out validation set from the official train set. (a) RGB$\rightarrow$IR, (b) SAR$\rightarrow$EO, (c) SAR$\rightarrow$IR, (d) SAR$\rightarrow$RGB. Each panel shows input, generated output, and ground-truth target. EarthBridge preserves structural integrity while synthesizing high-fidelity textures across EO, IR, and SAR modalities.}
    \label{fig:teaser}
\end{center}
}{}{}
\makeatother

\def\paperID{*****} 
\def\confName{CVPR}
\def\confYear{2026}

\title{EarthBridge: A Solution for 4th Multi-modal Aerial View Image Challenge Translation 
Track}

\author{%
\begin{tabular}{@{}c@{}}
\begin{tabular}[t]{@{}c@{\hspace{1em}}c@{\hspace{1em}}c@{}}
Zhenyuan Chen$^{*}$ &
Yuanshen Guan$^{*}$ &
Feng Zhang \\
\small Zhejiang University &
\small \begin{tabular}[t]{@{}c@{}}University of Science and Technology\\of China\end{tabular} &
\small Zhejiang University \\
{\tt\small bili\_sakura@zju.edu.cn} &
{\tt\small guanys@mail.ustc.edu.cn} &
{\tt\small zfcarnation@zju.edu.cn}
\end{tabular}\\[0.45em]
{\small $^{*}$Equal contribution. Corresponding author: {\tt zfcarnation@zju.edu.cn}.}
\end{tabular}
}
\begin{document}

\maketitle

\begin{abstract}
Cross-modal image-to-image translation among Electro-Optical (EO), Infrared (IR), and Synthetic Aperture Radar (SAR) sensors is essential for comprehensive multi-modal aerial-view analysis. However, translating between these modalities is notoriously difficult due to their distinct electromagnetic signatures and geometric characteristics. This paper presents \textbf{EarthBridge}, a high-fidelity framework developed for the 4th Multi-modal Aerial View Image Challenge -- Translation (MAVIC-T). Our main approach builds on \textbf{Diffusion Bridge Implicit Models (DBIM)}, using their non-Markovian bridge factorization over discretized timesteps, Karras-weighted training, and \emph{booting noise} at the source endpoint to stabilize sampling while encoding cross-modal ambiguity. Deterministic DBIM updates support very low function-evaluation counts when validation favors them; we choose per-task inference depths to optimize the official composite score, from a handful of steps on RGB$\rightarrow$IR to much larger budgets where metrics improve with additional steps. For SAR$\rightarrow$IR we submit \textbf{Contrastive Unpaired Translation (CUT)}, which outperforms diffusion bridges on our held-out validation under the same protocol. A channel-concatenated UNet denoiser implements the diffusion tracks. We evaluate on all four tasks (SAR$\rightarrow$EO, SAR$\rightarrow$RGB, SAR$\rightarrow$IR, RGB$\rightarrow$IR) and achieve a composite score of 0.38, ranking second on the MAVIC-T leaderboard.
\textbf{Code:} \url{https://github.com/Bili-Sakura/EarthBridge-Preview}.
\end{abstract}

\section{Introduction}
\label{sec:intro}

The Multi-modal Aerial View Image Challenge for cross-domain image translation~\cite{lowMultimodalAerialView2024,lowMultimodalAerialView2023,bowald3rdMultimodalAerial2025} is a leading challenge held in conjunction with the Perception Beyond the Visible Spectrum (PBVS) workshop from 2023 to 2025.
Imagery from multiple sensing modalities---EO, IR, and SAR---provides complementary views of the same scene, yet data coverage limitations and modality-specific constraints often prevent practitioners from fully exploiting these sources together.

For instance, SAR can offer all-weather sensing and penetrates many atmospheric obstructions~\cite{moreira2013tutorial}, but its imagery is harder to interpret~\cite{majumder2020deep,huangClassificationLargeScaleHighResolution2021} and paired or large-scale public corpora remain relatively limited compared with optical streams~\cite{lowMultimodalAerialView2024,bowald3rdMultimodalAerial2025}.
Conversely, widely available EO imagery is constrained by lighting, clouds, and atmospheric visibility~\cite{lowMultimodalAerialView2024}.
Domain translation mitigates this mismatch by mapping one modality to another, supporting cross-modal synthesis and interoperability~\cite{isolaImageToImageTranslationConditional2017}.
The MAVIC-T track focuses on RGB$\rightarrow$IR, SAR$\rightarrow$EO, SAR$\rightarrow$IR, and SAR$\rightarrow$RGB translation to grow usable data volume and diversity and to benchmark conditioned image synthesis for multi-modal aerial analysis.
\cref{sec:related_work} reviews prior translation and bridge methods and, for each theme, states what remains open under this kind of fixed, multi-task challenge evaluation and how EarthBridge is positioned.

SAR images possess unique attributes that present challenges for both human observers and vision artificial intelligence (AI) models to interpret, owing to their electromagnetic (EM) characteristics~\cite{majumder2020deep,huangClassificationLargeScaleHighResolution2021,huangDeepSARNetLearning2020,huangHDECTFAUnsupervisedLearning2021,huangPhysicallyExplainableCNN2022}.

In this paper, we propose \textbf{EarthBridge}, a multi-method framework for cross-modal aerial image translation. Our primary approach follows DBIM~\cite{zhengDiffusionBridgeImplicit2025} as published---including its non-Markovian bridge factorization, deterministic sampler, and booting-noise rule at $t{=}T$---and applies it to MAVIC-T; we do not alter the core DBIM sampling mathematics. In parallel, we explore CUT~\cite{parkContrastiveLearningUnpaired2020} as a powerful contrastive-learning-based alternative. Unlike standard conditional diffusion models that diffuse targets to pure noise and rely on classifier-free guidance or simple concatenation for the reverse path, EarthBridge implements a genuine stochastic diffusion bridge. By directly constraining the forward generative trajectory between the source modality $x_T$ and target modality $x_0$, and parameterizing the UNet~\cite{ronnebergerUNetConvolutionalNetworks2015} denoiser to solve the bridge marginals under Karras-weighted scalings~\cite{karras2022elucidating}, EarthBridge (DBIM) achieves high-fidelity translation across disparate modalities. We utilize pixel-level modeling throughout both approaches to preserve the full spectral and spatial resolution of the remote sensing data.

Our findings indicate that the deterministic DBIM sampling significantly improves inference speed, achieving high-quality results in as few as 5 sampling steps in RGB$\rightarrow$IR. Qualitatively, EarthBridge demonstrates a superior ability to reconstruct complex urban layouts and fine-grained textures in challenging tasks such as SAR-to-RGB, effectively bridging the domain gap. Quantitatively, our solution achieves a composite task score of 0.269 for the SAR$\rightarrow$EO task and 0.38 overall, ultimately placing second in the MAVIC-T challenge leaderboard.

Our main contributions are summarized as follows:
\begin{itemize}[nosep, leftmargin=*]
    \item We propose \textbf{EarthBridge}, a multi-method framework for paired cross-modal aerial image translation incorporating both DBIM and CUT, providing a robust comparison between diffusion-based and contrastive-learning-based approaches.
    \item We develop a specialized training protocol using Karras-weighted bridge scalings and pixel-level modeling, optimized for the high dynamic range and resolution requirements of multimodal remote sensing data.
    \item We conduct comprehensive experiments on all four MAVIC-T tasks (SAR$\rightarrow$EO, SAR$\rightarrow$RGB, SAR$\rightarrow$IR, RGB$\rightarrow$IR), reporting competitive results and providing detailed quantitative and qualitative analysis.
\end{itemize}
    
\section{Related Work}
\label{sec:related_work}

Cross-modal translation for aerial and satellite sensing supports fusion and downstream analysis, yet papers vary widely in supervision (paired vs.\ unpaired), generative family (GAN vs.\ diffusion), resolution, and evaluation protocol. We organize prior work into general image-to-image translation, diffusion bridge models, and RS-focused translators. In each case we highlight what is typically missing for \emph{paired, challenge-benchmarked} multi-track aerial translation and how EarthBridge relates.

\paragraph{Image-to-image and cross-modality translation.}
Paired conditional translation was popularized by Pix2Pix~\cite{isolaImageToImageTranslationConditional2017}; unpaired regimes rely on cycle consistency and adversarial alignment (CycleGAN~\cite{zhuUnpairedImageToImageTranslation2017}). Contrastive patch losses (CUT~\cite{parkContrastiveLearningUnpaired2020}) improve structure preservation when perfect alignment is noisy. Recent lines explore zero-shot and one-step diffusion-style translators~\cite{parmarZeroshotImagetoImageTranslation2023,parmarOneStepImageTranslation2024}. Cross-spectral colorization and transfer~\cite{oliveiraProbabilisticApproachColor2015,suarezInfraRedImageryColorization2018,zhangColorfulImageColorization2016,majumder2020deep} show that grayscale or NIR inputs can be mapped to plausible RGB, but targets and metrics are usually photographic rather than georeferenced mosaics at mixed native resolutions. What remains open for MAVIC-T-style settings is a \emph{single} leaderboard over four aligned aerial tracks ($256^2$ and $1024^2$) with identical composite scoring, where one must choose per-task objectives without hand-tuning to unrelated benchmarks. EarthBridge therefore routes SAR$\rightarrow$IR through CUT when official test metrics favor a feed-forward map, while retaining DBIM bridges elsewhere (\cref{sec:method,sec:experiments}).

\paragraph{Diffusion bridge models.}
Standard diffusion generates from a noise prior; \emph{bridge} formulations instead constrain paths between two endpoint distributions, which matches paired modality translation. Variance-preserving bridges appear in DDBM~\cite{zhouDenoisingDiffusionBridge2024}; related ideas include BBDM and bidirectional variants~\cite{Li_2023_CVPR,xueBiBBDMBidirectionalImage2025}, image-to-image Schr\"odinger bridges and matching objectives~\cite{liuI2SBImagetoImageSchrodinger2023,shiDiffusionSchrodingerBridge2023}, dual and direct implicit bridges with data consistency~\cite{suDualDiffusionImplicit2023a,chungDirectDiffusionBridge2023}, consistency-trained bridges~\cite{heConsistencyDiffusionBridge2024}, DBIM for fast implicit sampling~\cite{zhengDiffusionBridgeImplicit2025}, latent bridge matching~\cite{chadebecLBMLatentBridge2025a}, deterministic Brownian-bridge approximators~\cite{xiaoDeterministicImagetoImageTranslation2025a}, and adaptive handling of domain shift~\cite{wangAdaptiveDomainShift2026a}. Much of this literature emphasizes algorithmic transport and sample efficiency on natural-image benchmarks. Remote-sensing practice, by contrast, often still adopts noise-conditional DDPM/DDIM-style paths with task-specific refiners, and published bridge papers are rarely scored on the same multi-task aerial challenge. EarthBridge adopts DBIM~\cite{zhengDiffusionBridgeImplicit2025} \emph{as published}---without changing its sampler---and pairs it with ADM-style UNets, mixed-resolution training, and official MAVIC-T evaluation, including test-set sweeps over step count and bridge baselines (\cref{sec:experiments,sec:val_ablations}).

\paragraph{RS image-to-image translation.}
SAR-to-optical synthesis has moved from GANs with hierarchical or ViT-enhanced generators~\cite{wangSARtoOpticalImageTranslation2022,huGanbasedSAROptical2022,heDOGANDINOBasedOpticalPriorDriven2025,zhaoHVTcGANHybridVision2025,yangS3OILSemiSupervisedSARtoOptical2025} toward diffusion models with spatial--frequency conditioning~\cite{qinConditionalDiffusionModel2024,baiConditionalDiffusionSAR2024}, distilled one-step pipelines~\cite{qinEfficientEndtoEndDiffusion2025}, layout-aware robust refinement~\cite{zhaoRLIDMRobustLayoutBased2025}, domain-shift adaptation~\cite{wangAdaptiveDomainShift2026a}, unified Any2Any DiT-style backbones~\cite{chenAny2AnyUnifiedArbitrary2026}, and asymmetry-aware fusion designs such as CSHNet~\cite{yangCSHNetNovelInformation2026}. These contributions advance fidelity on their chosen geographies and modality pairs, but they seldom report all of RGB$\rightarrow$IR, SAR$\rightarrow$EO/IR/RGB under one fixed server-side metric at both low and megapixel resolutions. EarthBridge targets exactly that MAVIC-T setting: we cover every official task, use crop-based $512{\rightarrow}1024$ training for the high-resolution optical tracks following common compute-saving practice (cf.\ \cref{par:resolution512}), and document where diffusion bridges win or lose relative to CUT using competition test scores (\cref{tab:main_results,tab:val_test_sweeps}).

\section{Preliminary}
\label{sec:preliminary}

\paragraph{Problem Formulation}
Let $\mathcal{X} \subset \mathbb{R}^{C_s \times H \times W}$ denote the source domain and $\mathcal{Y} \subset \mathbb{R}^{C_t \times H \times W}$ denote the target domain, where $C_s$ and $C_t$ are the number of channels in the source and target modalities, respectively. Given a set of spatially-aligned training pairs $\{(x_i, y_i)\}_{i=1}^{N}$ with $x_i \in \mathcal{X}$ and $y_i \in \mathcal{Y}$, the goal of cross-modal image-to-image translation is to learn a mapping $G: \mathcal{X} \rightarrow \mathcal{Y}$ that minimizes a composite distance between the generated image $\hat{y} = G(x)$ and the ground-truth target $y$. In the context of the MAVIC-T challenge, four translation tasks are defined (see \cref{tab:dataset_tasks}): SAR$\rightarrow$EO ($1 \!\rightarrow\! 1$ channel, $256 \times 256$), SAR$\rightarrow$RGB ($1 \!\rightarrow\! 3$, $1024 \times 1024$), SAR$\rightarrow$IR ($1 \!\rightarrow\! 1$, $1024 \times 1024$), and RGB$\rightarrow$IR ($3 \!\rightarrow\! 1$, $1024 \times 1024$).

\paragraph{Evaluation Metric}
The MAVIC-T challenge evaluates submissions using a composite score derived from three complementary metrics: LPIPS~\cite{zhangUnreasonableEffectivenessDeep2018} (Learned Perceptual Image Patch Similarity, based on VGG-16) for perceptual fidelity, FID~\cite{heuselGANsTrainedTwo2017} (Fr\'{e}chet Inception Distance, based on InceptionV3) for distributional similarity, and L1 norm for pixel-level structural accuracy. These are combined into a task score:
\begin{equation}
    S_{\text{task}} = \frac{\tfrac{2}{\pi}\arctan(\text{FID}) + \text{LPIPS} + \text{L1}}{3}\,,
    \label{eq:task_score}
\end{equation}
where the $\arctan$ normalization maps FID into the $[0, 1]$ range to balance its contribution against the other two metrics. The overall score is the average of the four task scores, with a penalty of $+1$ added for each unattempted domain, encouraging participants to develop generalizable solutions across all modality pairs.

\paragraph{Notation.}
Throughout this paper, we use the abbreviations SDE (stochastic differential equation), ODE (ordinary differential equation), SNR (signal-to-noise ratio), NFE (number of function evaluations of the denoiser), and VP (variance-preserving) schedule.

\section{EarthBridge}
\label{sec:method}

We propose EarthBridge, built on DBIM~\cite{zhengDiffusionBridgeImplicit2025} for high-fidelity cross-modal translation, and CUT~\cite{parkContrastiveLearningUnpaired2020} for SAR$\rightarrow$IR where it outperforms diffusion bridges on the MAVIC-T \textbf{test-set} competition metrics (\cref{tab:val_sar_ir_rgb}).

\subsection{DBIM}
\label{sec:dbim_formalism}

Diffusion models have recently revolutionized generative modeling, yet they are fundamentally designed to transport complex distributions to a standard Gaussian prior. Adapting them to translate between two arbitrary distributions, such as heterogeneous remote sensing modalities, requires conditioning the diffusion process on a fixed endpoint. This is achieved via Doob's $h$-transform~\cite{doob1984classical,rogers2000diffusions}.

\paragraph{Stochastic Bridges via $h$-transform.}
Given a base diffusion process $dx_t = f(x_t, t)dt + g(t)dw_t$, the conditioned SDE that almost surely reaches a prescribed target $x_0 = y$ at $t=0$ starting from a source $x_T = x$ at $t=T$ is given by:
\begin{align}
    dx_t &= f(x_t, t)dt + g(t)dw_t \nonumber \\
    &\quad + g(t)^2 \nabla_{x_t} \log p(x_0 = y \mid x_t)dt,
    \label{eq:bridge_sde}
\end{align}
where $p(x_0 = y \mid x_t)$ is the transition density of the original process. When the endpoints are fixed, the resulting conditioned process is called a \emph{diffusion bridge}~\cite{sarkka2019applied,heng2021simulating,delyon2006simulation,schauer2017guided,peluchettinon,liu2022let}, which forms the foundational mechanism for transporting one state to another. In the context of cross-modal translation, $x$ represents the source modality (e.g., SAR) and $y$ the target modality (e.g., optical). This formulation ensures that the generative trajectory is directly constrained by both endpoints, providing a more principled framework for paired translation than standard conditional diffusion.

\begin{figure*}[t]
    \centering
    \includegraphics[page=2,width=\linewidth,trim=0 54pt 0 65pt,clip]{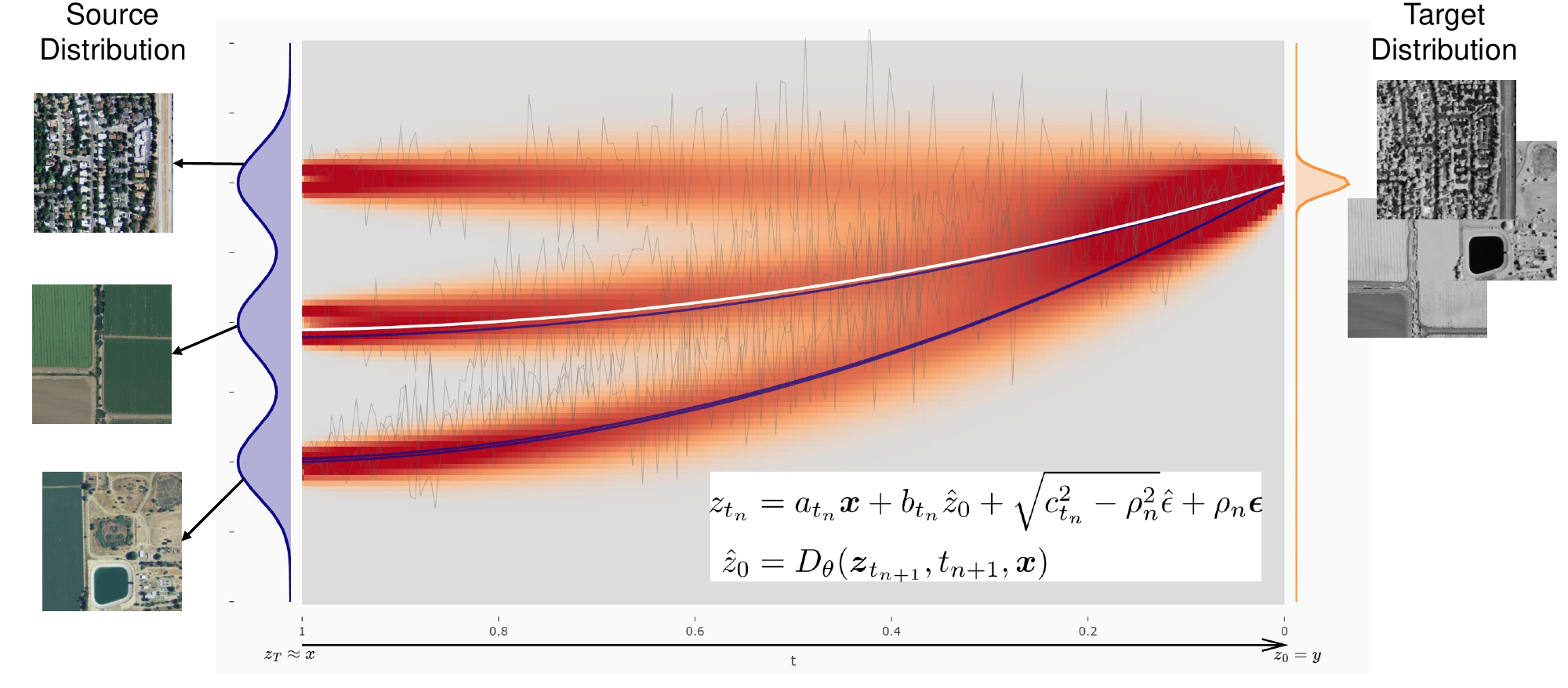}
    \caption{Distribution evolution of the diffusion bridge. The source distribution at $\bm{z}_T=\bm{x}$ (left) converges through time $t$ to the target distribution at $\bm{z}_0=\bm{y}$ (right), governed by the DBIM update rule and denoiser $D_\theta$.}
    \label{fig:bridge_evolution}
\end{figure*}

\paragraph{DDBM.}
\label{sec:vp_schedule}
DDBM~\cite{zhouDenoisingDiffusionBridge2024} provide a tractable realization of Eq.~\ref{eq:bridge_sde}. Under a variance-preserving (VP) schedule, the forward bridge process constructs intermediate states $z_t$ as:
\begin{equation}
    z_t = a_t x + b_t y + \sigma_t \epsilon, \quad \epsilon \sim \mathcal{N}(0, I),
    \label{eq:bridge_forward}
\end{equation}
where $a_t, b_t, \sigma_t$ are schedule-dependent coefficients. While DDBM excels in high-fidelity generation, its inference typically relies on simulating Eq.~\ref{eq:bridge_sde} via computationally expensive numerical solvers (e.g., Heun or stochastic samplers), often requiring hundreds of function evaluations (NFE) for optimal results. In this work, we adopt the more efficient \emph{implicit} formulation to accelerate this process. \cref{fig:bridge_evolution} illustrates the distribution evolution from source to target.

\subsection{EarthBridge: Non-Markovian DBIM}
\label{sec:earthbridge_nonmarkovian}

\begin{figure*}[t]
    \centering
    \includegraphics[page=4,width=\linewidth]{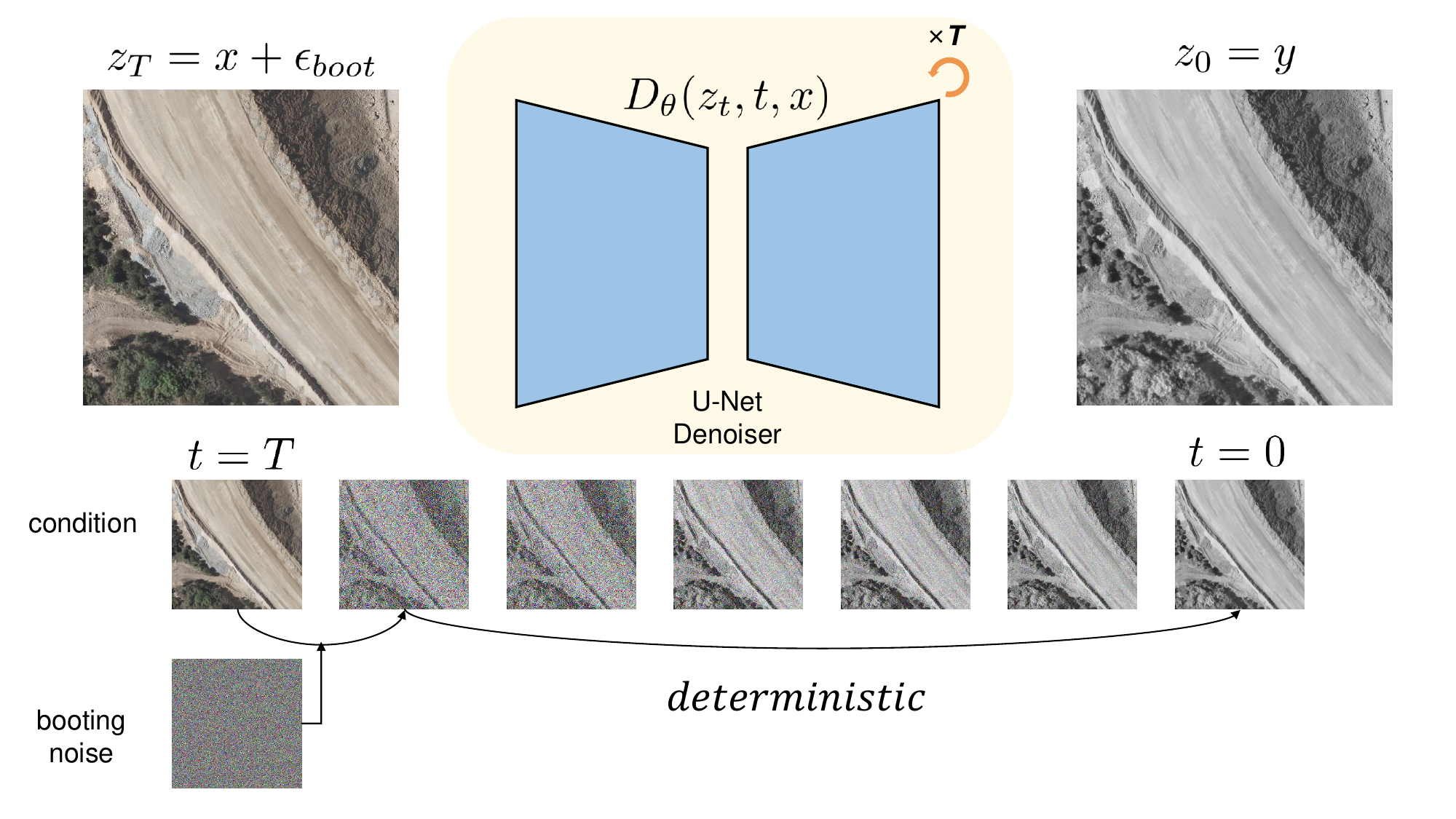}
    \caption{Deterministic DBIM sampling. The noisy input $\bm{z}_T=\bm{x}+\bm{\epsilon}_{\text{boot}}$ is progressively denoised by $D_\theta(\bm{z}_t, t, \bm{x})$ to produce the clean output $\bm{z}_0=\bm{y}$. The step sequence shows conditioning, booting noise injection, and gradual refinement from $t=T$ to $t=0$.}
    \label{fig:denoising_process}
\end{figure*}

\begin{figure*}[t]
    \centering
    \includegraphics[page=3,width=\linewidth]{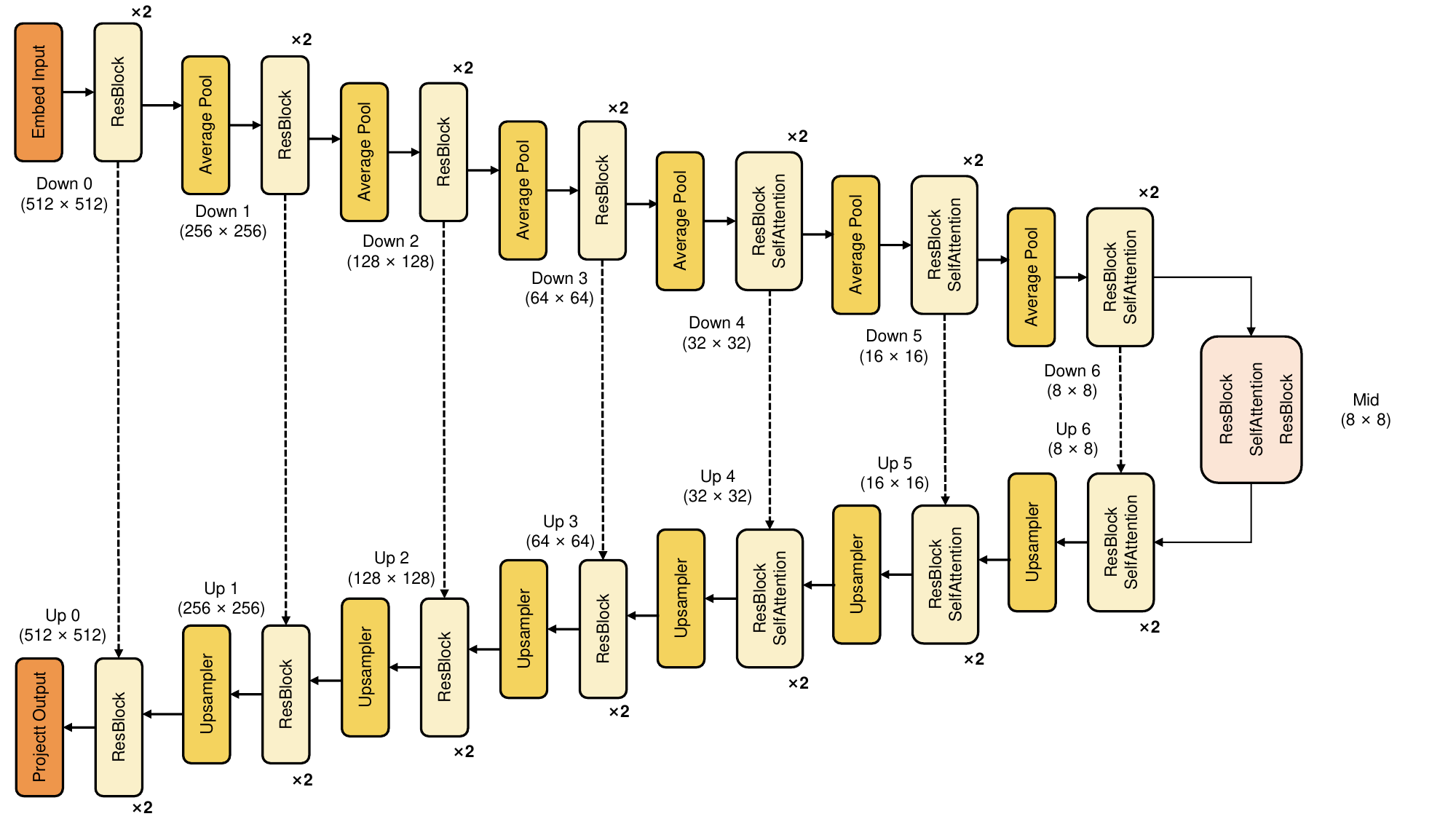}
    \caption{UNet denoiser architecture for 1024px tasks (SAR$\rightarrow$RGB, RGB$\rightarrow$IR). The encoder downsamples via ResBlocks and average pooling; the decoder upsamples with skip connections. Source $\bm{x}$ is channel-concatenated with $\bm{z}_t$; the output is $\hat{\bm{z}}_0$. We train at 512$\times$512 and infer at full resolution, following precedent that diffusion models trained at 512px can still synthesize faithfully at 1024px at inference time~\cite{rombachHighResolutionImageSynthesis2022}, which greatly reduces training cost versus training directly at 1024px.}
    \label{fig:unet_architecture}
\end{figure*}

EarthBridge uses DBIM~\cite{zhengDiffusionBridgeImplicit2025} \emph{as defined in the original paper}: the non-Markovian $\rho$-chain, marginal consistency with the VP bridge of Zhou et al.~\cite{zhouDenoisingDiffusionBridge2024}, the update in \cref{eq:dbim_update}, and booting noise at the endpoint (\cref{par:booting}) are the same sampling technology---not a modified variant. Our contribution is to instantiate this bridge for MAVIC-T (tasks, data, UNet, resolution, training details). Relative to simulating the DDBM stochastic bridge~\cite{zhouDenoisingDiffusionBridge2024}, DBIM's implicit discretization supports much faster inference at comparable fidelity. \cref{fig:denoising_process,fig:unet_architecture} show the sampling diagram and our ADM-style UNet~\cite{dhariwalDiffusionModelsBeat2021}.

\paragraph{Non-Markovian Diffusion Bridges.}
Following DBIM~\cite{zhengDiffusionBridgeImplicit2025} (rather than the continuous-time Markovian bridge in DDBM~\cite{zhouDenoisingDiffusionBridge2024}), we use the same non-Markovian factorization over discretized timesteps $0 = t_0 < t_1 < \dots < t_N = T$, controlled by a variance parameter $\rho \in \mathbb{R}^{N-1}$. The UNet denoiser is trained to predict the bridge marginals governed by the schedule-dependent coefficients $(a_t, b_t, c_t)$, where the source image $\bm{x}$ is supplied to the network via channel concatenation. This concatenation is strictly an architectural mechanism to inject the mandatory endpoint parameter $\bm{x}$ into the SDE formulation, not an ad-hoc conditioning trick akin to standard image-to-image models. The joint probability distribution $q^{(\rho)}(\bm{z}_{t_{0:N-1}} \mid \bm{x})$ is factorized through a chain of conditional distributions:
\begin{equation}
    q^{(\rho)}(\bm{z}_{t_{0:N-1}} \mid \bm{x}) = q_0(\bm{z}_{t_0}) \prod_{n=1}^{N-1} q^{(\rho)}(\bm{z}_{t_n} \mid \bm{z}_0, \bm{z}_{t_{n+1}}, \bm{x}),
    \label{eq:joint_dist}
\end{equation}
where $q_0$ is the target modality distribution and each step $q^{(\rho)}$ is defined as:
\begin{equation}
    \mathcal{N}\big( a_{t_n} \bm{x} + b_{t_n} \bm{z}_0 + \sqrt{c_{t_n}^2 - \rho_n^2}\, \bm{\eta}_{n+1},\, \rho_n^2 \bm{I} \big),
\end{equation}
with $\bm{\eta}_{n+1} = (\bm{z}_{t_{n+1}} - a_{t_{n+1}} \bm{x} - b_{t_{n+1}} \bm{z}_0) / c_{t_{n+1}}$. Here, $a_t, b_t, c_t$ are known coefficients determined by the VP log-SNR schedule (\cref{sec:vp_schedule}). Under this construction, the marginal distributions $q^{(\rho)}(\bm{z}_{t_n} \mid \bm{x})$ are preserved to be consistent with the continuous-time forward bridge (\cref{eq:bridge_forward}). The variance parameter $\rho$ controls the level of stochasticity in the sampling process, ranging from a purely Markovian stochastic process to a fully deterministic implicit model.

\paragraph{Reverse Generative Process and Sampling.}
The generative process is initialized at the source image $\bm{z}_T = \bm{x}$ and iteratively samples $\bm{z}_{t_n}$ from $\bm{z}_{t_{n+1}}$ via a parameterized data predictor $D_\theta(\bm{z}_{t_{n+1}}, t_{n+1}, \bm{x})$. The core updating rule is given by:
\begin{equation}
    \bm{z}_{t_n} = a_{t_n} \bm{x} + b_{t_n} \hat{\bm{z}}_0 + \sqrt{c_{t_n}^2 - \rho_n^2} \hat{\bm{\epsilon}} + \rho_n \bm{\epsilon},
    \label{eq:dbim_update}
\end{equation}
where $\hat{\bm{z}}_0 = D_\theta(\bm{z}_{t_{n+1}}, t_{n+1}, \bm{x})$ is the clean data prediction and $\hat{\bm{\epsilon}}$ is the predicted noise at time $t_{n+1}$. Setting $\rho_n = 0$ for all steps yields the deterministic \emph{DBIM} sampler, which can be viewed as an Euler discretization of a probability flow ODE specifically tailored for diffusion bridges.

\paragraph{Booting Noise.}
\label{par:booting}
This is DBIM's prescribed fix for the $t{=}T$ endpoint singularity~\cite{zhengDiffusionBridgeImplicit2025}, which we adopt unchanged: since $c_{t_N} = 0$, the expression for $\hat{\bm{\epsilon}}$ is ill-defined unless the first reverse step is stochastic. DBIM sets $\rho_{N-1} = c_{t_{N-1}}$, injecting standard Gaussian \emph{booting noise} $\bm{\epsilon}_{\text{boot}}$. We do not ablate booting; removing it would restore the singularity discussed in~\cite{zhengDiffusionBridgeImplicit2025}. In remote sensing translation, that stochasticity also provides a practical latent degree of freedom for one-to-many mappings (e.g., several plausible infrared images for one SAR input).

\paragraph{Channel Handling for Mismatched Modalities.}
For tasks where source and target modalities have different channel counts (e.g., RGB$\rightarrow$IR or SAR$\rightarrow$RGB), we operate the bridge in a shared model channel space (typically 3 for RGB-related tasks). Single-band modalities are expanded to model channels by channel repetition before entering the UNet, ensuring the concatenation $[\bm{z}_t, \bm{x}]$ remains well-defined across all tasks. We use repetition for its minimal parameter count and stable optimization under our budget; alternative fusion designs that we did not fully evaluate are discussed in \cref{sec:scope_extensions}.

\subsection{CUT}
\label{sec:cut_formalism}

For SAR$\rightarrow$IR we use CUT~\cite{parkContrastiveLearningUnpaired2020} rather than DBIM. On the \textbf{official MAVIC-T test set}, CUT substantially improves the composite task score relative to both DBIM and DDBM under the same competition evaluation protocol (\cref{tab:val_sar_ir_rgb}); DBIM also incurs much higher inference cost at comparable quality. Unlike diffusion bridges, CUT is a feed-forward GAN that translates in one pass. It combines an adversarial loss with a patch-based contrastive loss to maintain structural correspondence between source and output. Architecture and hyperparameter details are in \cref{sec:suppl_architecture}.

\paragraph{PatchNCE loss and generator objective.}
CUT extracts multi-layer encoder features from both the source $\bm{x}$ and the generated image $\hat{\bm{y}} = G(\bm{x})$. A lightweight MLP samples spatial patches per layer, projects them to an embedding space, and applies L2 normalization. The PatchNCE (InfoNCE~\cite{oordRepresentationLearningContrastive2019}) loss encourages that the query embedding from a patch in $\hat{\bm{y}}$ is closer to the key embedding from the corresponding patch in $\bm{x}$ than to embeddings from other patches. Formally, for query $q$ and key $k$ at the same spatial location, the loss uses $\ell_{\text{NCE}} = -\log \frac{\exp(q^\top k / \tau)}{\exp(q^\top k / \tau) + \sum_{k'} \exp(q^\top k' / \tau)}$, with temperature $\tau$ and negatives $k'$ from other patches. This contrastive objective preserves structural layout across the translation. The generator is trained to minimize $\mathcal{L}_G = \lambda_{\text{GAN}} \mathcal{L}_{\text{GAN}} + \lambda_{\text{NCE}} \mathcal{L}_{\text{NCE}}$, where $\mathcal{L}_{\text{GAN}}$ is the LSGAN~\cite{maoLeastSquaresGenerative2017} objective (the generator encourages the discriminator to classify $G(\bm{x})$ as real) and $\mathcal{L}_{\text{NCE}}$ is the per-layer PatchNCE loss averaged over the selected encoder layers. \Cref{sec:objectives} states how this objective sits alongside DBIM training in EarthBridge.

\section{Experiments}
\label{sec:experiments}

\subsection{Dataset and Preprocessing}
\label{sec:dataset}

We use the MAVIC-T 2025 dataset from the challenge organizers, with imagery from UNICORN, USGS EROS HRO, and UMBRA across four regions: Bingham, Centerfield, Manhattan, and UC Davis. We do not use any external datasets beyond the official challenge data. \cref{tab:dataset_tasks} summarizes the four translation tasks. SAR$\rightarrow$EO operates at 256$\times$256; the other three tasks use 1024$\times$1024. For SAR$\rightarrow$IR and SAR$\rightarrow$RGB, we spatially align SAR scenes with optical mosaics before resampling. We also create crop-augmented training sets for the 1024$\times$1024 tasks to increase data diversity.

\begin{table}[t]
  \caption{Task specifications and training data for the MAVIC-T challenge. Crop augmentation uses 1024$\times$1024 sliding windows with stride 512 (50\% overlap); SAR$\rightarrow$optical pairs are georeference-aligned.}
  \label{tab:dataset_tasks}
  \centering
  \small
  \setlength{\tabcolsep}{3pt}
  \begin{tabular*}{\columnwidth}{@{\extracolsep{\fill}}lcccc@{}}
    \toprule
    Task & Resol. & Ch. & Samp.\ (orig.) & Samp.\ (aug.) \\
    \midrule
    SAR$\rightarrow$EO  & 256$^2$   & 1$\rightarrow$1 & 89,411 & -- \\
    SAR$\rightarrow$RGB & 1024$^2$ & 1$\rightarrow$3 & 10,576 & 75,703 \\
    SAR$\rightarrow$IR  & 1024$^2$ & 1$\rightarrow$1 & 10,576 & 74,317 \\
    RGB$\rightarrow$IR  & 1024$^2$ & 3$\rightarrow$1 & 2,272  & 7,388  \\
    \bottomrule
  \end{tabular*}
\end{table}

\subsection{Per-track models and training objectives}
\label{sec:objectives}

\paragraph{Architectures and task routing.}
For DBIM-based tasks (SAR$\rightarrow$EO, SAR$\rightarrow$RGB, RGB$\rightarrow$IR), we use a UNet denoiser that predicts the clean target from the noisy sample conditioned on timestep and source image. For SAR$\rightarrow$IR, we adopt CUT~\cite{parkContrastiveLearningUnpaired2020} because it attains a much better \textbf{test-set} competition task score than DBIM/DDBM at lower latency (\cref{tab:val_sar_ir_rgb}). Per-task architectural and hyperparameter choices are summarized in \cref{tab:suppl_arch_dbim,tab:suppl_arch_cut}.

\paragraph{Training objectives.}
\textbf{DBIM.} We train the denoiser with a weighted mean squared error between the prediction and the clean target:
\begin{equation}
    \mathcal{L}_{\text{denoise}} = \mathbb{E}_{(\bm{x},\bm{y}),\, t,\, \epsilon} \left[ w(t) \cdot \| D_\theta(\bm{z}_t, t, \bm{x}) - \bm{y} \|_2^2 \right],
    \label{eq:denoising_loss}
\end{equation}
where $w(t)$ is the Karras bridge weighting. Diffusion time $t$ is sampled from a Karras inverse-$\rho$ distribution; inference uses deterministic DBIM sampling.

\textbf{CUT.} For SAR$\rightarrow$IR only, training follows the adversarial + PatchNCE generator objective in \cref{sec:cut_formalism}; we do not use Eq.~\ref{eq:denoising_loss} on that track.

Full technical details on architecture, schedule, and per-task configurations are in \cref{sec:suppl_architecture}.

\subsection{Quantitative Results}
\label{sec:quantitative}

\begin{table*}[t]
  \caption{Main results on the MAVIC-T evaluation set. $\downarrow$ lower is better. Time is per-sample inference latency; SAR$\to$EO is run at 256px with large batch size, and its latency is averaged.}
  \label{tab:main_results}
  \centering
  \small
  \setlength{\tabcolsep}{4pt}
  \renewcommand{\arraystretch}{1.05}
  \begin{tabular*}{\textwidth}{@{\extracolsep{\fill}}llcccccc@{}}
    \toprule
    Track & Method & FID$\downarrow$ & LPIPS$\downarrow$ & L1$\downarrow$ & Score$\downarrow$ & NFE & Time \\
    \midrule
    SAR$\to$EO  & DBIM & 0.22 & 0.50 & 0.08 & 0.27 & 500 & 0.42\,s \\
    SAR$\to$RGB & DBIM & 0.88 & 0.64 & 0.21 & 0.58 & 1000 & 160\,s \\
    SAR$\to$IR  & CUT & 0.65 & 0.60 & 0.15 & 0.46 & 1 & 0.47\,s \\
    RGB$\to$IR  & DBIM & 0.36 & 0.15 & 0.09 & 0.20 & 5 & 1.5\,s \\
    \bottomrule
  \end{tabular*}
\end{table*}

\begin{table*}[t]
  \caption{Leaderboard comparison. Per-task scores (SAR$\to$EO, SAR$\to$RGB, SAR$\to$IR, RGB$\to$IR) from the challenge evaluation.}
  \label{tab:leaderboard_comparison}
  \centering
  \small
  \setlength{\tabcolsep}{4pt}
  \renewcommand{\arraystretch}{1.05}
  \begin{tabular*}{\textwidth}{@{\extracolsep{\fill}}lcccccc@{}}
    \toprule
    Rank & Participant & SAR$\to$EO & SAR$\to$RGB & SAR$\to$IR & RGB$\to$IR & Combined \\
    \midrule
    1st & HNU-VPAI & \textbf{0.11} & \textbf{0.50} & \underline{0.49} & \textbf{0.20} & \textbf{0.32} \\
    2nd & EarthBridge (\textbf{Ours}) & \underline{0.27} & \underline{0.58} & \textbf{0.46} & \textbf{0.20} & \underline{0.38} \\
    \bottomrule
  \end{tabular*}
\end{table*}

\cref{tab:main_results} presents the quantitative performance comparison between EarthBridge and the CUT baseline across the challenge tasks. \cref{tab:leaderboard_comparison} compares our second-place submission with the first-place result (HNU-VPAI); note that lower scores indicate better performance (the official leaderboard display was initially reversed). The composite task score provides a balanced evaluation of perceptual, distributional, and structural fidelity. EarthBridge demonstrates consistent performance across both low- and high-resolution tasks.

\paragraph{Validation and NFE selection.}
The per-task NFE in \cref{tab:main_results} can read as inconsistent with ``few-step'' DBIM (e.g., $500$--$1000$ vs.\ $5$ for RGB$\rightarrow$IR): DBIM supports low-NFE sampling everywhere, but we tune $N$ per task under the challenge composite score. Full \textbf{test-set} competition sweeps---NFE for RGB$\rightarrow$IR and SAR$\rightarrow$EO, CUT vs.\ DBIM/DDBM for SAR$\rightarrow$IR, and SAR$\rightarrow$RGB baselines (same server metrics as the leaderboard; FID uses normalized $\mathrm{atan}$)---are in \cref{sec:val_ablations,tab:val_test_sweeps}.

\subsection{Qualitative Results}
\label{sec:qualitative}

\begin{figure}[t]
    \centering
    \includegraphics[page=5,width=\columnwidth,trim=150pt 0 160pt 0,clip]{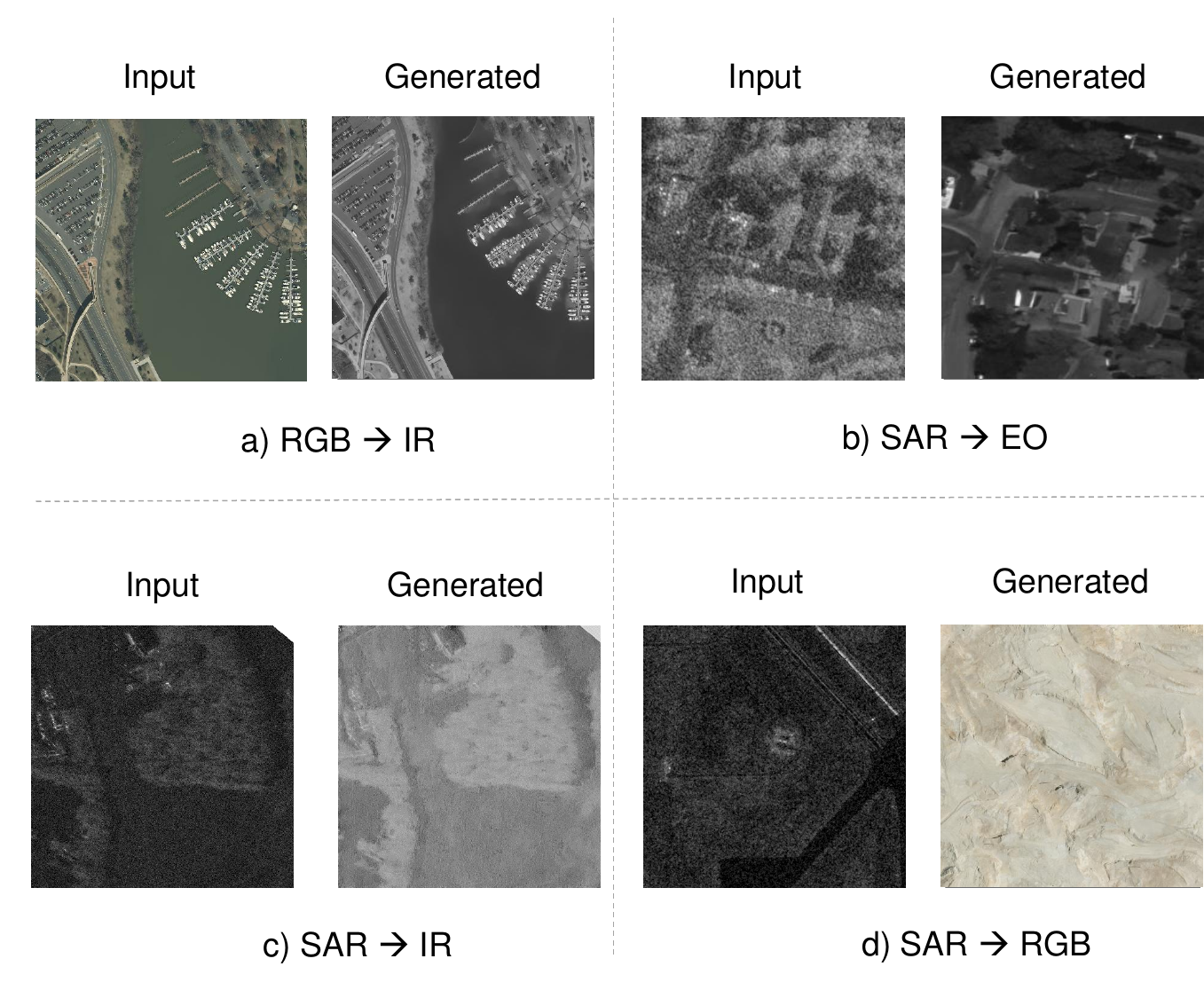}
    \caption{Qualitative results across all four MAVIC-T translation tasks. Each row corresponds to one task (SAR$\rightarrow$EO, SAR$\rightarrow$RGB, SAR$\rightarrow$IR, RGB$\rightarrow$IR). Columns show: source image, EarthBridge output, and ground-truth target. Our model preserves structural layout from the source while synthesizing faithful target-modality textures across both 256$\times$256 and 1024$\times$1024 resolutions.}
    \label{fig:qualitative}
\end{figure}

\cref{fig:qualitative} shows qualitative examples across all four tasks. EarthBridge preserves fine-grained structural information from the source modality while accurately synthesizing target textures, demonstrating consistent performance across both low- and high-resolution settings.

\section{Implementation Details}
\label{sec:suppl_impl}

This section provides the comprehensive technical details of the EarthBridge architecture, training hyperparameters, and multi-modal refinement modules. We use the Prodigy optimizer~\cite{mishchenkoProdigyExpeditiouslyAdaptive2024} for all training. Training employs mixed precision in BF16; inference is also run in BF16.

\subsection{Architecture, Sampling, and Scheduler Settings}
\label{sec:suppl_architecture}

\paragraph{DBIM.}
The denoiser follows the ADM-style UNet described in \cref{sec:earthbridge_nonmarkovian} (see also~\cite{dhariwalDiffusionModelsBeat2021,ronnebergerUNetConvolutionalNetworks2015}). Source conditioning is via channel-wise concatenation $[\bm{z}_t, \bm{x}]$. For SAR$\rightarrow$RGB, we use REPA~\cite{yuRepresentationAlignmentGeneration2025} in our implementation, which relieves the loss spike issue in the early training steps. Sampling follows the deterministic DBIM formulation~\cite{zhengDiffusionBridgeImplicit2025} with task-dependent inference steps (\cref{tab:main_results}) and Gaussian booting noise $\bm{\epsilon}_{\text{boot}}$ at $t=T$. \Cref{tab:suppl_arch_dbim} summarizes per-task DBIM architectural and training configurations, including the VP log-SNR schedule and Karras diffusion-time sampling.

\paragraph{Resolution (512 train, 1024 infer).}
\label{par:resolution512}
For native 1024$\times$1024 challenge outputs (SAR$\rightarrow$RGB, SAR$\rightarrow$IR, RGB$\rightarrow$IR), we train UNet/CUT backbones on 512$\times$512 crops and run inference at full 1024$\times$1024. Stable Diffusion~\cite{rombachHighResolutionImageSynthesis2022} demonstrates that diffusion models trained at 512px can still synthesize well at 1024px at inference, greatly reducing training cost versus native 1024px training; we apply the same \emph{train lower, infer higher} resolution strategy here. Under our competition timeline and compute budget we did not run a full 1024px training stage; further gains may be possible by warm-starting from our checkpoints and fine-tuning for a small number of epochs at 1024px resolution.

\paragraph{CUT.}
The CUT formulation (PatchNCE loss, training objective) is described in \cref{sec:cut_formalism}. Our implementation follows the original architecture~\cite{parkContrastiveLearningUnpaired2020} with three core components: a ResNet-based generator, a PatchGAN~\cite{isolaImageToImageTranslationConditional2017} discriminator, and a PatchSampleMLP for patch-wise contrastive learning. \Cref{tab:suppl_arch_cut} lists CUT settings for SAR$\to$IR.

\begin{table}[t]
  \caption{DBIM per-task architectural and training settings (SAR$\to$EO, SAR$\to$RGB, RGB$\to$IR).}
  \label{tab:suppl_arch_dbim}
  \centering
  \footnotesize
  \setlength{\tabcolsep}{2pt}
  \renewcommand{\arraystretch}{1.03}
  \begin{tabularx}{\columnwidth}{@{}>{\raggedright\arraybackslash}Xccc@{}}
    \toprule
    Configuration & SAR$\to$EO & SAR$\to$RGB & RGB$\to$IR \\
    \midrule
    Parameters & $\sim$20M & $\sim$120M & $\sim$120M \\
    VP schedule ($\beta_d$, $\beta_{\min}$) & 2.0, 0.1 & 2.0, 0.1 & 2.0, 0.1 \\
    Karras $\rho$ & 7 & 7 & 7 \\
    Base channels & 64 & 96 & 128 \\
    Channel mult. & (1,2,3,4) & (1,1,2,2,4,4) & (1,1,2,2,4,4) \\
    Attn.\ resol. & N/A & 32, 16, 8 & 32, 16, 8 \\
    Train resol. & $256^2$ & $512^2$ & $512^2$ \\
    Epochs & 55 & 24 & 86 \\
    \bottomrule
  \end{tabularx}
\end{table}

\begin{table}[t]
  \caption{CUT architectural and training settings for SAR$\to$IR (ResNet-9-block generator, 4-layer PatchGAN discriminator).}
  \label{tab:suppl_arch_cut}
  \centering
  \footnotesize
  \setlength{\tabcolsep}{2pt}
  \renewcommand{\arraystretch}{1.03}
  \begin{tabularx}{\columnwidth}{@{}>{\raggedright\arraybackslash}Xc@{}}
    \toprule
    Configuration & SAR$\to$IR \\
    \midrule
    Params (total) & $\sim$225M \\
    \quad Gen.; Disc. & $\sim$180M; $\sim$45M \\
    Gen.\ (ResNet blk.) & 9 \\
    Discriminator & 4-layer PatchGAN \\
    Filters (ngf, ndf) & 256 \\
    GAN objective & LSGAN \\
    $\tau$; $\lambda_{\mathrm{GAN}}$; $\lambda_{\mathrm{NCE}}$ & 0.1; 0.5; 1.0 \\
    Identity NCE & off \\
    Train resol. & $512^2$ \\
    Epochs & 11 \\
    \bottomrule
  \end{tabularx}
\end{table}

\section{Scope, Limitations, and Optional Extensions}
\label{sec:scope_extensions}

This section summarizes limitations of our MAVIC-T submission and outlines natural extensions.

\paragraph{Scope of the reported results.}
Our quantitative evaluation in \cref{tab:main_results,tab:val_test_sweeps} covers DBIM for SAR$\rightarrow$EO, SAR$\rightarrow$RGB, and RGB$\rightarrow$IR; CUT for SAR$\rightarrow$IR; and task-dependent NFE under the official composite metric.

\paragraph{Limitations.}
For 1024px tasks we train at 512px and infer at 1024px (\cref{par:resolution512}); we did not add a second stage of native-1024 fine-tuning under the competition budget. In channel-mismatched settings we use channel repetition as the default bridge interface and do not report a full fusion ablation against alternatives (\cref{sec:earthbridge_nonmarkovian}).

\paragraph{Future work.}
Promising directions include short native-resolution fine-tuning after 512px training and replacing channel repetition with learned stems or modality-aware fusion modules; we leave systematic study of these variants to future work.

\hypersetup{pageanchor=true}

\clearpage
{
    \small
    \setlength{\parskip}{0pt}
    \bibliographystyle{ieeenat_fullname}
    \bibliography{main}
}

\appendix
\clearpage
\hypersetup{pageanchor=false}
\setcounter{table}{0}
\renewcommand{\thetable}{S\arabic{table}}

\section{Competition test set: NFE selection and \texorpdfstring{SAR$\rightarrow$IR}{SAR2IR} model choice}
\label{sec:val_ablations}

\cref{tab:main_results} in the main paper reports per-task NFE that may appear inconsistent with ``few-step'' DBIM sampling (e.g., $500$--$1000$ for some tracks vs.\ $5$ for RGB$\rightarrow$IR). DBIM is \emph{capable} of very low-NFE sampling everywhere; the discrepancy reflects a \emph{task-dependent} choice of step count under the challenge's composite metric, not a limitation of the sampler. \cref{tab:val_test_sweeps} reports \textbf{official MAVIC-T competition scores on the challenge test set}---the same evaluation pipeline and composite metric as the leaderboard (FID uses normalized $\mathrm{atan}$), with each row corresponding to a distinct evaluated configuration submitted to the server. \textbf{Light-blue rows} indicate the best composite score within each task block.

\begin{table*}[t]
  \caption{MAVIC-T \textbf{test-set} competition scores (all tasks). \textbf{Light-blue rows} mark the best composite score per task block (full-row fill). $\downarrow$ lower is better. DDBM uses a numerical bridge sampler with the listed steps. CUT rows list generator capacity in the Method column and use NFE${=}1$ (single feed-forward pass). \emph{Nano Banana Pro} is an external image-generation baseline; we omit NFE (---) when the system does not expose an equivalent step count.}
  \label{tab:val_test_sweeps}
  \label{tab:val_nfe_rgb_eo}
  \label{tab:val_sar_ir_rgb}
  \centering
  \definecolor{lightbluerow}{RGB}{226,239,255}
  \footnotesize
  \setlength{\tabcolsep}{3pt}
  \renewcommand{\arraystretch}{1.03}
  \begin{tabularx}{\textwidth}{@{}ll>{\centering\arraybackslash}X>{\centering\arraybackslash}X>{\centering\arraybackslash}X>{\centering\arraybackslash}X>{\centering\arraybackslash}X@{}}
    \toprule
    Task & Method & NFE & FID$\downarrow$ & LPIPS$\downarrow$ & L1$\downarrow$ & Score$\downarrow$ \\
    \midrule
    \multicolumn{7}{@{}l}{\textit{RGB$\rightarrow$IR} (NFE sweep)} \\
    \midrule
    RGB$\rightarrow$IR & DBIM & 1 & 0.786 & 0.293 & 0.093 & 0.390 \\
    RGB$\rightarrow$IR & DBIM & 2 & 0.342 & 0.195 & 0.091 & 0.209 \\
    \cellcolor{lightbluerow}RGB$\rightarrow$IR & \cellcolor{lightbluerow}DBIM & \cellcolor{lightbluerow}5 & \cellcolor{lightbluerow}0.357 & \cellcolor{lightbluerow}0.151 & \cellcolor{lightbluerow}0.090 & \cellcolor{lightbluerow}\textbf{0.200} \\
    RGB$\rightarrow$IR & DBIM & 10 & 0.380 & 0.150 & 0.090 & 0.206 \\
    RGB$\rightarrow$IR & DBIM & 20 & 0.601 & 0.468 & 0.076 & 0.382 \\
    RGB$\rightarrow$IR & DBIM & 100 & 0.478 & 0.188 & 0.108 & 0.258 \\
    RGB$\rightarrow$IR & DBIM & 1000 & 0.474 & 0.225 & 0.110 & 0.270 \\
    RGB$\rightarrow$IR & DDBM & 100 & 0.709 & 0.366 & 0.154 & 0.410 \\
    \midrule
    \multicolumn{7}{@{}l}{\textit{SAR$\rightarrow$EO} (NFE sweep)} \\
    \midrule
    SAR$\rightarrow$EO & DBIM & 20 & 0.737 & 0.464 & 0.074 & 0.425 \\
    SAR$\rightarrow$EO & DBIM & 100 & 0.451 & 0.477 & 0.077 & 0.335 \\
    \cellcolor{lightbluerow}SAR$\rightarrow$EO & \cellcolor{lightbluerow}DBIM & \cellcolor{lightbluerow}500 & \cellcolor{lightbluerow}0.222 & \cellcolor{lightbluerow}0.504 & \cellcolor{lightbluerow}0.080 & \cellcolor{lightbluerow}\textbf{0.269} \\
    SAR$\rightarrow$EO & DBIM & 1000 & 0.233 & 0.511 & 0.079 & 0.274 \\
    \midrule
    \multicolumn{7}{@{}l}{\textit{SAR$\rightarrow$IR} (CUT vs.\ diffusion bridges)} \\
    \midrule
    \cellcolor{lightbluerow}SAR$\rightarrow$IR & \cellcolor{lightbluerow}CUT (huge) & \cellcolor{lightbluerow}1 & \cellcolor{lightbluerow}0.645 & \cellcolor{lightbluerow}0.603 & \cellcolor{lightbluerow}0.145 & \cellcolor{lightbluerow}\textbf{0.464} \\
    SAR$\rightarrow$IR & CUT (large) & 1 & 0.825 & 0.610 & 0.157 & 0.531 \\
    SAR$\rightarrow$IR & DDBM & 100 & 0.874 & 0.616 & 0.244 & 0.578 \\
    SAR$\rightarrow$IR & DBIM & 100 & 0.956 & 0.706 & 0.256 & 0.640 \\
    SAR$\rightarrow$IR & DBIM & 1000 & 0.958 & 0.630 & 0.194 & 0.594 \\
    \midrule
    \multicolumn{7}{@{}l}{\textit{SAR$\rightarrow$RGB} (NFE vs.\ baselines)} \\
    \midrule
    \cellcolor{lightbluerow}SAR$\rightarrow$RGB & \cellcolor{lightbluerow}DBIM & \cellcolor{lightbluerow}1000 & \cellcolor{lightbluerow}0.878 & \cellcolor{lightbluerow}0.636 & \cellcolor{lightbluerow}0.214 & \cellcolor{lightbluerow}\textbf{0.576} \\
    SAR$\rightarrow$RGB & DDBM & 100 & 0.894 & 0.656 & 0.236 & 0.595 \\
    SAR$\rightarrow$RGB & Nano Banana Pro & --- & 0.957 & 0.740 & 0.208 & 0.635 \\
    \bottomrule
  \end{tabularx}
\end{table*}

For RGB$\rightarrow$IR, the composite score is best at $N{=}5$ steps and degrades for both very small and very large $N$ (\cref{tab:val_test_sweeps}, RGB$\rightarrow$IR block); this matches the $N{=}5$ entry in \cref{tab:main_results}. For SAR$\rightarrow$EO, scores improve sharply between $N{=}100$ and $N{=}500$ and are near-optimal at $500$--$1000$ steps (\cref{tab:val_test_sweeps}, SAR$\rightarrow$EO block), which motivates the high-NFE settings used for that track despite DBIM's support for fewer steps. For SAR$\rightarrow$RGB, DBIM at $N{=}1000$ outperforms DDBM at $N{=}100$ and the Nano Banana Pro image-generation baseline on the \textbf{test set} (\cref{tab:val_test_sweeps}, SAR$\rightarrow$RGB block). For SAR$\rightarrow$IR, CUT clearly beats DBIM/DDBM on task score on the \textbf{test set} (\cref{tab:val_test_sweeps}, SAR$\rightarrow$IR block); we therefore submit CUT for that track.

\paragraph{Relation to the main paper.}
\cref{tab:val_test_sweeps} reports official MAVIC-T \textbf{test-set} competition scores under alternate step counts and model choices; it complements \cref{tab:main_results} by showing how per-task NFE was selected under the composite metric and how CUT compares to diffusion-bridge baselines for SAR$\rightarrow$IR. \cref{sec:scope_extensions} states training-resolution and channel-handling limitations of our submission.

\hypersetup{pageanchor=true}

\end{document}